\documentclass[11pt,a4paper]{article}
\usepackage[hyperref]{naaclhlt2019}
\usepackage{times}
\usepackage{latexsym}
\usepackage{CJKutf8}
\usepackage{graphicx}
\usepackage{booktabs}
\usepackage{array}

\usepackage{url}

\aclfinalcopy 

\title{INS: An Interactive Chinese News Synthesis System}

\author{Hui Liu \and Wentao Qin \and Xiaojun Wan \\
	Institute of Computer Science and Technology, Peking University \\
	The MOE Key Laboratory of Computational Linguistics, Peking University \\
	{\tt \{xinkeliuhui,qinwentao,wanxiaojun\}@pku.edu.cn} \\}

\date{}

\begin{document}
\begin{CJK*}{UTF8}{gbsn}
\maketitle
\begin{abstract}
  Nowadays, we are surrounded by more and more online news articles. Tens or hundreds of news articles need to be read if we wish to explore a hot news event or topic. So it is of vital importance to automatically synthesize a batch of news articles related to the event or topic into a new synthesis article (or overview article) for reader's convenience. It is so challenging to make news synthesis fully automatic that there is no successful solution by now. In this paper, we put forward a novel Interactive News Synthesis system (i.e. INS), which can help generate news overview articles automatically or by interacting with users. More importantly, INS can serve as a tool for editors to help them finish their jobs. In our experiments, INS performs well on both topic representation and synthesis article generation. A user study also demonstrates the usefulness and users' satisfaction with the INS tool. A demo video is available at \url{https://youtu.be/7ItteKW3GEk}. 
\end{abstract}

\section{INTRODUCTION}
In the last decade, news websites and apps become more and more popular, which can provide us an extremely large volume of news articles. Even for a single news event, there are usually tens or hundreds of related news articles published online. In order to have a complete image of a news event, we have to look through all of the related news articles, which is very time-consuming and inefficient. With a large amount of time spent, what we get is just fragmented information scattered in different news articles.

Ideally, if there exists an overview article about an event, we will fully and efficiently understand the event by reading it. However, such overview articles are not easy to write, even for professional editors, and the writing of them is very time-consuming. Though existing multi-document summarization systems can produce short summaries by selecting several representative sentences, they cannot produce long overview articles with good structure and high quality. So it is of vital importance to design a system to help editors or users to efficiently synthesize a batch of related news articles into a long news overview article, either in a fully automatic way or in a semi-automatic way.

To achieve the above goal, we put forward a novel Interactive News Synthesis system (i.e. INS), which can help generate Chinese news overview articles automatically or by interacting with users. Given a news event or topic, INS first crawls news articles from major Chinese news websites, detects different subtopics and represents them with easy-to-understand labels. Afterward, a span of text for each subtopic will be generated and the news synthesis article will be organized accordingly. It is noteworthy that INS can interact with users in different stages.

We automatically evaluate the key component of the INS system, i.e., subtopic detection and representation, and evaluation results demonstrate its efficacy. Human evaluation is employed and a user study is performed to demonstrate the usefulness and users' satisfaction of the INS tool. 

\section{Related Work}
One of the related fields is document summarization. The methods can be divided into extractive methods \cite{gillick2009scalable,lin2010multi,berg2011jointly,sipos2012large,woodsend2012multiple,wan2014ctsum,nallapati2017summarunner,ren2017leveraging} and abstractive methods \cite{rush2015neural,nallapati2016abstractive,tan2017abstractive}. 

There are several pilot studies on producing long articles from a batch of news articles or web pages\cite{yao2011autopedia,zhang2017towards,liu2018generating}. However, the generated overview articles do not have good structures and there are no interaction functions.

There are some attempts of adding interaction functions into the traditional document summarization tasks \cite{jones2002interactive,leuski2003ineats}. However, the above work focuses on producing short summaries and the generation of long news overview articles is more challenging. Moreover, in the above work, the keyphrases to represent salient information are extracted based on some heuristic rules or simple clues, and they are usually not good subtopic representations. 

\section{System Overview and User Interaction}

Our INS system aims to produce a long overview article for a specific news event by synthesizing a number of existing Chinese news articles. It consists of the following components: 

\textbf{News Fetcher}: According to the input news event, INS first crawls relevant news articles by using a popular Chinese news search engine (e.g. Baidu News), and extracts the title and body text for each news article. The number of news articles can be set by users and it is set to 100 by default.  

\textbf{Subtopic Finder}: This component aims to discover the subtopics in the news articles and represent them with some informative labels. We extract n-grams that meet some demands as candidate labels and leverage a regression model to predict a score for each candidate label. Top 20 labels will be chosen and merged, after which each label represents a specific subtopic.

\textbf{Article Synthesizer}: This component aims to produce a span of text with moderate length for each subtopic, and then assemble the texts of selected subtopics to form the final news overview article. We first split the original news articles into coherent text blocks. The text blocks are then matched with the subtopic label and ranked by using the topic-sensitive TextRank algorithm and the MMR redundancy removal method. We select one or several top-ranked text blocks to describe each subtopic. Finally, all the selected text blocks are assembled to form the final overview article, with the subtopic labels used as the subtitles of the blocks.  

\begin{figure*}[htb]
	\centering
	\includegraphics[width=0.9\textwidth]{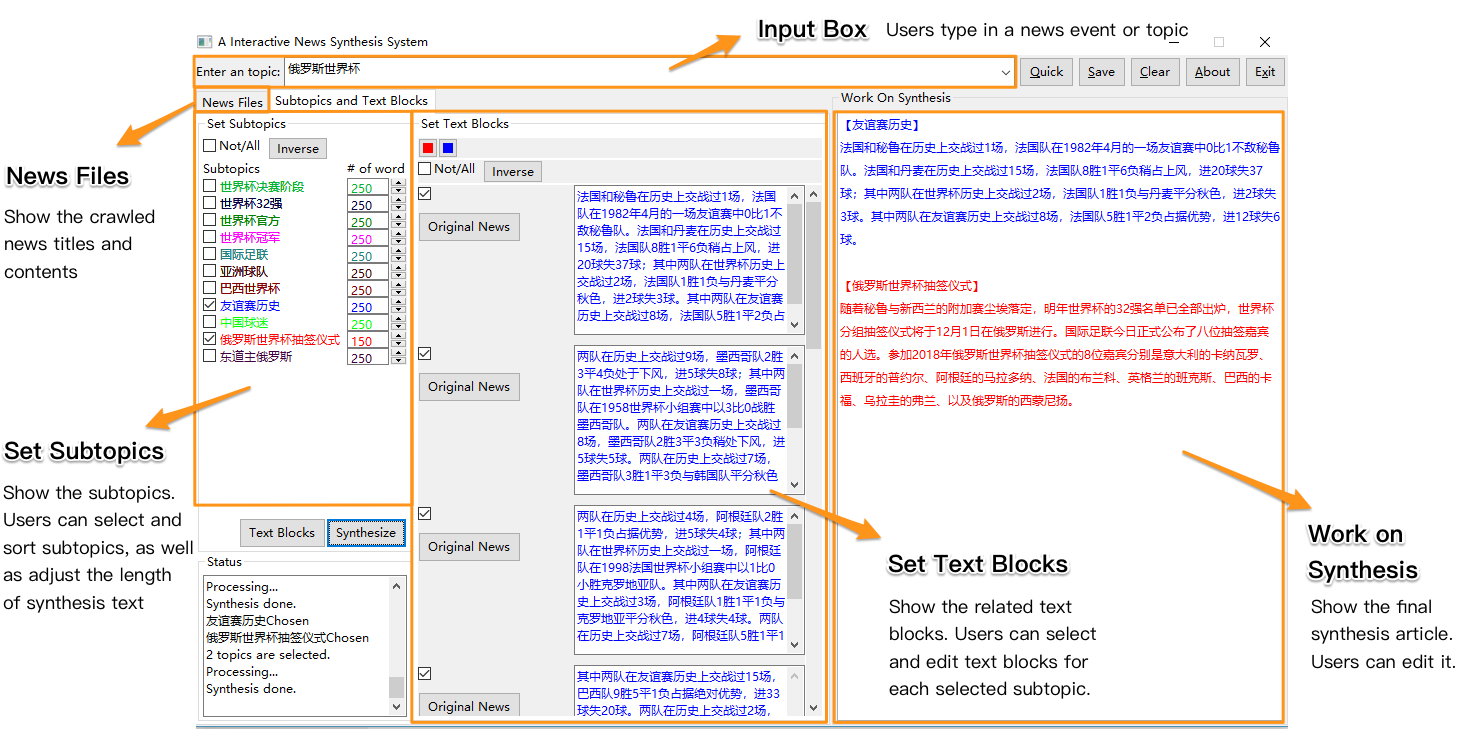}
	\caption{The GUI of our INS system. \small{(Select two subtopics: ‘俄罗斯世界杯抽签仪式’ (The Russian World Cup draw ceremony) and ‘友谊赛历史’ (History of friendly match), and change the order of the two subtopics. Select the text blocks for each subtopic and make some edits on the text blocks. Then click on the synthesize button to obtain the final synthesis article for editing.)}}
	\label{fig:GUI}
\end{figure*}

The graphic user interface of INS is shown in Figure \ref{fig:GUI} with the input topic of ‘俄罗斯世界杯/2018 FIFA World Cup Russia’.

INS can interact with users in different stages: Users can choose and re-order some of the labels according to their preferences; Users can choose and edit the text blocks for each chosen subtopic; Users can edit on the final news synthesis result to get a more excellent article. Note that the interactions are optional. For `lazy' users, INS can generate the final synthesis article with just one click. 

\section{Subtopic Finder}
There are some existing unsupervised methods for detecting subtopics from documents, for example, text clustering and topic models. They have some drawbacks in common: (1) The number of subtopics (clusters) needs to be manually set beforehand and inappropriate subtopic number will affect the final results significantly. (2) It is hard to represent each subtopic, which is very important for users' understanding and selection of the subtopics.

Different from the above methods which first detect the subtopics or clusters and then extract labels for each subtopic, we decide to directly find subtopic labels by using supervised learning, and the labels are informative and easy to understand. We first extract candidate labels from the news articles and then use a regression model to assign a rating score to each candidate. After that, we merge the top labels to obtain the final subtopic labels.

\textbf{Candidate Label Extraction}: We extract n-grams from the original news articles as our candidate labels, where n ranges from 1 to 3. We employ pyltp for Chinese word segmentation and POS tagging. An n-gram will be a candidate if it meets the following requirements: (1) Its term frequency is higher than a min-count threshold. We set min-count to 25 for unigrams and 10 for bigrams and trigrams. (2) It is not a substring of the input topic name, which is too general to be a subtopic label. (3) It does not include time words and adverbs. (4) A candidate label of unigram is limited to only nouns and verbs. We use these rules to filter out the n-grams that are not suitable for representing subtopics.

\textbf{Label Score Prediction}: We adopt regression models to predict a score for each candidate label. We choose 12 features to describe each label and each feature is expected to indicate whether a candidate is good or not from some aspect. These features include Term Frequency-Inverse Document Frequency, Document Frequency, Number of words in the label, Number of Chinese characters in the label, Intra-Cluster Similarity, Cluster Entropy, Independence Entropy, Frequency of the label in news titles, Syntactic continuity, Number of nouns in the label, and topic model information.

In the training data, each candidate label is manually assigned with a score of 0\textasciitilde3, and a higher score indicates a better label. We then train regression models for label score prediction. We explore and compare different regression models, and finally choose support vector regression (SVR).

\textbf{Label Merging}: After label score prediction, the candidate labels are sorted by their predicted scores and the top 20 labels are reserved. We will merge similar labels according to the following rules: (1) If two n-grams share a common part which is more than one word, we will merge them into one label. (2) If one label is a substring of another, we only reserve the label with a higher score. (3) If two labels' cosine similarity is greater than a threshold (0.65 in this study), we only reserve the label with a higher score. After merged, labels are shown to users and each label stands for a subtopic.

\section{Article Synthesizer}

In this step, we need to produce a span of text with moderate length to describe each subtopic and then combine all the texts into the final synthesis article. We first segment the news articles into coherent text blocks and then rank the relevant text blocks for each subtopic. One or more salient text blocks can be chosen to describe the corresponding subtopic.

\textbf{Text Segmentation}: Most of the methods for document summarization use sentences as the basic unit, which is not appropriate for long article generation. Synthesis articles using the sentence as the basic unit tend to be fragmented and hard to read. As a result, we use the text block as our basic unit. A text block is a set of several continuous sentences and it can cover a relative complete idea. We use our proposed SenTiling algorithm \cite{zhang2017towards} to segment texts, which is a variant of Hearst's TextTiling algorithm \cite{hearst1997texttiling}. After text segmentation, news articles are divided into text blocks, which include 2.3 sentences on average.

\textbf{Text Block Ranking and Selection}: For each subtopic, INS system first selects candidate text blocks using exact match. A text block that contains a subtopic label is assigned to the subtopic. Then INS uses a topic sensitive TextRank algorithm to rank candidate text blocks for each subtopic where TextRank(\cite{mihalcea2004textrank}) is a typical graph-based ranking algorithm applied in document summarization. We build a graph with the candidate text blocks as vertexes and the similarity between text blocks as the weight of the edge. We can determine the importance $WS(V_i)$ of every vertex through a random walk with a restart. In our case, the restart probability of every vertex is set to the normalized similarity between each text block and the subtopic label so that the top text blocks are both important and relevant to the subtopic.

After ranking, the top text blocks may be similar and this leads to redundancy. We try to solve this problem by using the MMR criterion \cite{carbonell1998use}.

Now we have chosen the text blocks to describe each subtopic. Then we rearrange the text blocks according to the following strategies: (1) If two text blocks are extracted from one article, we sort them by the original order. (2) If two text blocks are extracted from different articles, we put the text blocks written earlier in the front. Because news event usually changes over time, this strategy can partly reconstruct the event process.

\textbf{Synthesis Article Construction}: Now we have labels for different subtopics and the corresponding text for each subtopic. INS then constructs the final news synthesis article. If users choose and rank the labels, INS will use their preferred order. If users choose and edit the text blocks, INS will use them instead of the default ones. Otherwise, INS chooses several labels with the highest scores and use the highly ranked text blocks of the subtopic labels for news synthesis. In this way, INS produces the final news synthesis article. Users can fix it to get their own article.

\section{EVALUATION}

\textbf{Data Set}:  We chose 20 news topics and crawled about 100 Chinese news articles for each news topic, 1969 articles in total. The news topics cover a wide range of fields, including politics(e.g., 萨德系统/THAAD Missile System), technology(e.g., 百度无人驾驶汽车/Baidu Self-driving Car), society(e.g., 江歌刘鑫案/The Case of Jiangge and Liuxin), entertainment(e.g., 绝地求生/PLAYERUNKNOWN'S BATTLEGROUND), and sports(e.g., 俄罗斯世界杯/2018 FIFA World Cup Russia).

In order to train and test the regression models for detecting subtopic labels, we extracted all the n-grams that meet the previously-mentioned requirements and got about 220 candidate labels for each topic on average. Then we tagged them manually, with each label assigned with a score from 0 to 3. A higher score means a better label. The majority of n-grams are assigned with 0 and the labels with nonzero scores account for 26.3\% in total. The labels with nonzero scores are considered acceptable subtopic labels. 

\textbf{Evaluation on Subtopic Labels}: We leverage all 12 features to train the SVR model, and use 20-fold cross-validation, with 19 news topics for training and the rest one for validation in turn. The values of P@5, P@10 and P@20 for SVR are 0.722, 0.693 and 0.619, respectively. The results indicate the majority of top 20 labels can represent subtopics. Other inappropriate labels can be filtered out by interaction with users.

\textbf{Evaluation on News Synthesis Articles}: There are no gold reference synthesis articles for each news topic, and it is also hard to manually write a few reference articles. Thus we choose to conduct a manual evaluation of the final news synthesis articles. We use the multi-document summarization methods implemented in PKUSUMSUM \cite{zhang2016pkusumsum} as baselines.

The summarization methods we choose include Lead, Coverage, Centroid, and TextRank. Centroid and TextRank can work on either the sentence unit (i.e., Centroid-sen, TextRank-sen) or the text block unit (i.e., Centroid-blk, TextRank-blk). Thus we have six baselines. The baselines are compared with INS that does not involve user interaction. Each baseline and INS generate a synthesis article with 1000 words for each news event and 10 news topics will be manually evaluated. We employ 16 Chinese college students as judges. Each student evaluates one or two news topics. We make sure each student judge all the 7 articles for each news topic and each news topic is evaluated by 3 students. The judges are asked to give a rating score between 1 and 6 from the following aspects: readability, structure, topic diversity, redundancy removal, and overall responsiveness. The average results are shown in Table \ref{tab:5}. From the table, we can see that INS achieves the best performance in every aspect, which proves the effectiveness of our INS system. We also perform pairwise t-tests when comparing our system with baselines, and find that INS significantly outperforms sentence-based methods in every aspect and block-based methods in almost all aspects (p-value $<$  0.05).

Note that in the above comparison, INS does not involve any user interaction. We believe after the interaction with users at different stages, the quality of the synthesis articles will be much improved. 

\begin{table}[htb]
	\centering
	\small
	\begin{tabular}{|p{1.8cm}<{\centering}|p{0.7cm}<{\centering}|p{0.7cm}<{\centering}|p{0.7cm}<{\centering}|p{0.7cm}<{\centering}|p{0.75cm}<{\centering}|}
		\hline
		Method & Read. & Struc. & Topic. & Redun. & Overall \\
		\hline
		Lead-sen & 4.214 & 3.500 & 4.107 & 3.964 & 3.964 \\
		\hline
		Coverage-sen & 3.250 & 2.714 & 4.179 & 3.179 & 3.107 \\
		\hline
		Centroid-sen & 4.179 & 3.714 & 3.964 & 4.250 & 4.036 \\
		\hline
		TextRank-sen & 3.929 & 3.464 & 3.929 & 3.929 & 3.786 \\
		\hline
		Centroid-blk & 5.036 & 4.857 & 4.179 & 4.571 & 4.929 \\
		\hline
		TextRank-blk & 4.893 & 4.536 & 4.393 & 4.643 & 4.679 \\
		\hline
		INS & \textbf{5.321} & \textbf{5.179} & \textbf{5.286} & \textbf{5.214} & \textbf{5.179} \\
		\hline
	\end{tabular}
	\caption{Manual evaluation results}
	\label{tab:5}
\end{table}


\textbf{User Study on INS}: We further performed a user study on INS by employing 10 users to experience it and give their judgments. 
The rating scores given by users on various aspects (e.g., usefulness, GUI, satisfaction, assistance) are generally high. This indicates that the INS system is useful and users are satisfied with the system. 

In addition to scoring, we asked them to give their comments on INS. The majority consider it as an excellent helper. They said subtopics are helpful to know about the given topic clearly, and INS can filter redundant information out. Besides, they think INS provide them with enough and useful interactions to produce synthesis articles. One user pointed out that the GUI of INS can be further improved. 

In the future, we will concentrate on improving the performance and beautifying the user interface. We also plan to deploy the system in several Chinese news media.  

\section*{Acknowledgment}

This work was supported by National Natural Science Foundation of
China (61772036) and Key Laboratory of Science, Technology and Standard
in Press Industry (Key Laboratory of Intelligent Press Media Technology).
We appreciate the anonymous reviewers for their helpful comments. Xiaojun Wan is the corresponding author.

\bibliography{naaclhlt2019}

\begin{thebibliography}{20}
\expandafter\ifx\csname natexlab\endcsname\relax\def\natexlab#1{#1}\fi

\bibitem[{Berg-Kirkpatrick et~al.(2011)Berg-Kirkpatrick, Gillick, and
  Klein}]{berg2011jointly}
Taylor Berg-Kirkpatrick, Dan Gillick, and Dan Klein. 2011.
\newblock Jointly learning to extract and compress.
\newblock In \emph{Proceedings of the 49th Annual Meeting of the Association
  for Computational Linguistics: Human Language Technologies-Volume 1}, pages
  481--490. Association for Computational Linguistics.

\bibitem[{Carbonell and Goldstein(1998)}]{carbonell1998use}
Jaime Carbonell and Jade Goldstein. 1998.
\newblock The use of mmr, diversity-based reranking for reordering documents
  and producing summaries.
\newblock In \emph{Proceedings of the 21st annual international ACM SIGIR
  conference on Research and development in information retrieval}, pages
  335--336. ACM.

\bibitem[{Gillick and Favre(2009)}]{gillick2009scalable}
Dan Gillick and Benoit Favre. 2009.
\newblock A scalable global model for summarization.
\newblock In \emph{Proceedings of the Workshop on Integer Linear Programming
  for Natural Langauge Processing}, pages 10--18. Association for Computational
  Linguistics.

\bibitem[{Hearst(1997)}]{hearst1997texttiling}
Marti~A Hearst. 1997.
\newblock Texttiling: Segmenting text into multi-paragraph subtopic passages.
\newblock \emph{Computational linguistics}, 23(1):33--64.

\bibitem[{Jones et~al.(2002)Jones, Lundy, and Paynter}]{jones2002interactive}
Steve Jones, Stephen Lundy, and Gordon~W Paynter. 2002.
\newblock Interactive document summarisation using automatically extracted
  keyphrases.
\newblock In \emph{System Sciences, 2002. HICSS. Proceedings of the 35th Annual
  Hawaii International Conference on}, pages 1160--1169. IEEE.

\bibitem[{Leuski et~al.(2003)Leuski, Lin, and Hovy}]{leuski2003ineats}
Anton Leuski, Chin-Yew Lin, and Eduard Hovy. 2003.
\newblock ineats: interactive multi-document summarization.
\newblock In \emph{Proceedings of the 41st Annual Meeting on Association for
  Computational Linguistics-Volume 2}, pages 125--128. Association for
  Computational Linguistics.

\bibitem[{Lin and Bilmes(2010)}]{lin2010multi}
Hui Lin and Jeff Bilmes. 2010.
\newblock Multi-document summarization via budgeted maximization of submodular
  functions.
\newblock In \emph{Human Language Technologies: The 2010 Annual Conference of
  the North American Chapter of the Association for Computational Linguistics},
  pages 912--920. Association for Computational Linguistics.

\bibitem[{Liu et~al.(2018)Liu, Saleh, Pot, Goodrich, Sepassi, Kaiser, and
  Shazeer}]{liu2018generating}
Peter~J Liu, Mohammad Saleh, Etienne Pot, Ben Goodrich, Ryan Sepassi, Lukasz
  Kaiser, and Noam Shazeer. 2018.
\newblock Generating wikipedia by summarizing long sequences.
\newblock \emph{arXiv preprint arXiv:1801.10198}.

\bibitem[{Mihalcea and Tarau(2004)}]{mihalcea2004textrank}
Rada Mihalcea and Paul Tarau. 2004.
\newblock Textrank: Bringing order into text.
\newblock In \emph{Proceedings of the 2004 conference on empirical methods in
  natural language processing}.

\bibitem[{Nallapati et~al.(2017)Nallapati, Zhai, and
  Zhou}]{nallapati2017summarunner}
Ramesh Nallapati, Feifei Zhai, and Bowen Zhou. 2017.
\newblock Summarunner: A recurrent neural network based sequence model for
  extractive summarization of documents.
\newblock In \emph{AAAI}, pages 3075--3081.

\bibitem[{Nallapati et~al.(2016)Nallapati, Zhou, Gulcehre, Xiang
  et~al.}]{nallapati2016abstractive}
Ramesh Nallapati, Bowen Zhou, Caglar Gulcehre, Bing Xiang, et~al. 2016.
\newblock Abstractive text summarization using sequence-to-sequence rnns and
  beyond.
\newblock \emph{arXiv preprint arXiv:1602.06023}.

\bibitem[{Ren et~al.(2017)Ren, Chen, Ren, Wei, Ma, and
  de~Rijke}]{ren2017leveraging}
Pengjie Ren, Zhumin Chen, Zhaochun Ren, Furu Wei, Jun Ma, and Maarten de~Rijke.
  2017.
\newblock Leveraging contextual sentence relations for extractive summarization
  using a neural attention model.
\newblock In \emph{Proceedings of the 40th International ACM SIGIR Conference
  on Research and Development in Information Retrieval}, pages 95--104. ACM.

\bibitem[{Rush et~al.(2015)Rush, Chopra, and Weston}]{rush2015neural}
Alexander~M Rush, Sumit Chopra, and Jason Weston. 2015.
\newblock A neural attention model for abstractive sentence summarization.
\newblock \emph{arXiv preprint arXiv:1509.00685}.

\bibitem[{Sipos et~al.(2012)Sipos, Shivaswamy, and Joachims}]{sipos2012large}
Ruben Sipos, Pannaga Shivaswamy, and Thorsten Joachims. 2012.
\newblock Large-margin learning of submodular summarization models.
\newblock In \emph{Proceedings of the 13th Conference of the European Chapter
  of the Association for Computational Linguistics}, pages 224--233.
  Association for Computational Linguistics.

\bibitem[{Tan et~al.(2017)Tan, Wan, and Xiao}]{tan2017abstractive}
Jiwei Tan, Xiaojun Wan, and Jianguo Xiao. 2017.
\newblock Abstractive document summarization with a graph-based attentional
  neural model.
\newblock In \emph{Proceedings of the 55th Annual Meeting of the Association
  for Computational Linguistics (Volume 1: Long Papers)}, volume~1, pages
  1171--1181.

\bibitem[{Wan and Zhang(2014)}]{wan2014ctsum}
Xiaojun Wan and Jianmin Zhang. 2014.
\newblock Ctsum: extracting more certain summaries for news articles.
\newblock In \emph{Proceedings of the 37th international ACM SIGIR conference
  on Research \& development in information retrieval}, pages 787--796. ACM.

\bibitem[{Woodsend and Lapata(2012)}]{woodsend2012multiple}
Kristian Woodsend and Mirella Lapata. 2012.
\newblock Multiple aspect summarization using integer linear programming.
\newblock In \emph{Proceedings of the 2012 Joint Conference on Empirical
  Methods in Natural Language Processing and Computational Natural Language
  Learning}, pages 233--243. Association for Computational Linguistics.

\bibitem[{Yao et~al.(2011)Yao, Jia, Shou, Feng, Zhou, and
  Liu}]{yao2011autopedia}
Conglei Yao, Xu~Jia, Sicong Shou, Shicong Feng, Feng Zhou, and Hongyan Liu.
  2011.
\newblock Autopedia: Automatic domain-independent wikipedia article generation.
\newblock In \emph{Proceedings of the 20th international conference companion
  on World wide web}, pages 161--162. ACM.

\bibitem[{Zhang and Wan(2017)}]{zhang2017towards}
Jianmin Zhang and Xiaojun Wan. 2017.
\newblock Towards automatic construction of news overview articles by news
  synthesis.
\newblock In \emph{Proceedings of the 2017 Conference on Empirical Methods in
  Natural Language Processing}, pages 2111--2116.

\bibitem[{Zhang et~al.(2016)Zhang, Wang, and Wan}]{zhang2016pkusumsum}
Jianmin Zhang, Tianming Wang, and Xiaojun Wan. 2016.
\newblock Pkusumsum: a java platform for multilingual document summarization.
\newblock In \emph{Proceedings of COLING 2016, the 26th International
  Conference on Computational Linguistics: System Demonstrations}, pages
  287--291.

\end{thebibliography}
\bibliographystyle{acl_natbib}

\end{CJK*}
\end{document}